\documentclass[11pt]{article}

\usepackage[preprint]{acl}

\usepackage{times}
\usepackage{latexsym}

\usepackage[T1]{fontenc}

\usepackage[utf8]{inputenc}

\usepackage{microtype}

\usepackage{inconsolata}

\usepackage{graphicx}

\usepackage{placeins}
\usepackage{amsmath}
\usepackage{amsfonts}
\usepackage{mathtools}
\usepackage{booktabs}
\usepackage{multirow}
\usepackage{float}
\title{Read the Room Before Your Model Generates}

\author{Muyao Kong$^{*\,\dagger}$ \\
  New York University \\
  \texttt{muyao.kong@nyu.edu} \\\And
  Zhihao Lin$^{*}$ \\
  Independent Researcher \\
  \texttt{zhihaolinsde@gmail.com} \\}

\begin{document}
\maketitle

\begingroup
\renewcommand{\thefootnote}{\fnsymbol{footnote}}
\footnotetext[1]{Equal contribution.}
\footnotetext[2]{Corresponding author.}
\endgroup

\begin{abstract}
Large language models (LLMs) generally rely on sampling temperature to regulate the amount of randomness in their outputs. Mainstream methods for selecting temperature typically involve fixed, heuristic global settings and context or token-level calibration, rather than a geometrically interpretable representation of the communicative scenario. In this work, we introduce a simple and interpretable temperature selection framework, the \textbf{Scenario-Actor}, which adapts decoding to the inferred situational characteristics. Scenario-Actor extracts scenario-relevant structures from Transformer hidden states and determines the temperature from scenario embedding, a structured geometric representation of the communicative setting. Empirical results show that on Llama-3.1-8B-Instruct, Scenario-Actor achieves strong goal-completion performance in interactive social scenarios, significant gains on open-ended instruction following, and competitive performance on multi-turn conversation compared to multiple baselines.\footnote{An interactive version of our paper is at \url{https://muyao-kong.github.io/scenario-actor/}.}
\end{abstract}

\section{Introduction}

Large language models (LLMs) generate text by repeatedly sampling from a conditional next-token distribution. One of the simplest yet most consequential hyperparameters in language generation is the sampling temperature, which controls the sharpness of the next-token probability distribution and, in turn, the rigor and creativity of model output. 

Existing systems typically set it globally or adapt it according to predictive uncertainty, context, or learned decoding objectives, which implicitly treat decoding primarily as a property of the model's predictive state \citep{dhuliawala2024adaptive,wang2020contextual}. But socially situated language generation exposes a different source of variation: the interaction itself. A prompt conveys not only semantics but also implies a communicative setting. For example, the conversation could be either an academic discussion or a friendly chitchat, whose desired responses are either rigorous or open-ended, formal or casual. The optimal temperature is then circumstantial, varying across scenarios.

\begin{figure}[htbp]
    \centering
    \includegraphics[width=0.95\columnwidth]{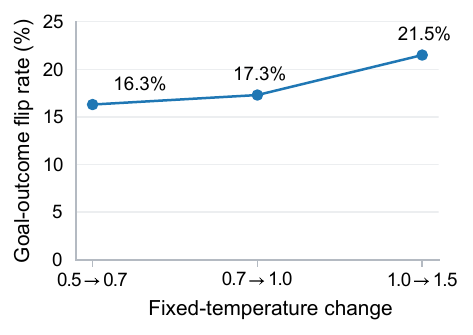}\\[2.5mm]
    \includegraphics[width=0.95\columnwidth]{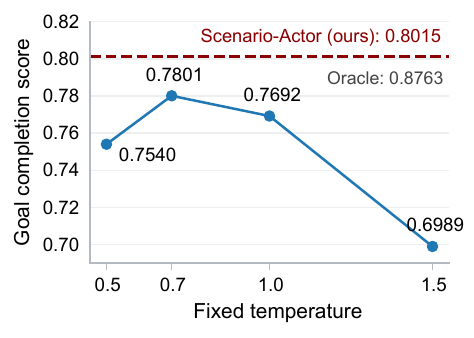}
    \caption{AgentSense goal-outcome flip rate (top) and goal completion score (bottom) by changing temperature.}
    \label{fig:scenario_actor}
\end{figure}

In our controlled experiment, we decode 245 AgentSense \citep{mou2025agentsense} social interactions with four fixed temperatures. We found that first, changing temperature substantially alters interaction quality, moving between mid-range temperatures flips 16 to 17\% of individual goal outcomes; second, globally strongest temperature setting does not dominate across interactions.

Such results show a major limitation of viewing temperature as a globally fixed hyperparameter, which motivates us to propose a way to control scenario-aware interaction and language generation. In this work, we propose \textbf{Scenario-Actor}, a framework for scenario-adaptive temperature selection. Scenario-Actor focuses on social interactions and open-ended instruction following, where the prompt provides enough evidence to infer a communicative scenario. We introduce a structured representation of the communicative setting extracted from context - the scenario embedding\footnote{In this work, we use the terms \textit{scenario vector} and \textit{scenario embedding} interchangeably.}, whose span defines the scenario space, and use it to derive the sampling temperature. Specifically, our method maps a representation from the $r$-th layer of the Transformer to an $m$-dimensional scenario space, where each input is represented by a scenario embedding. An $\operatorname{Actout}$ function then maps the geometry of this vector to a prompt-level sampling temperature.

Prior work in social psychology provides us with a theoretical support by introducing DIAMONDS \citep{rauthmann2014diamonds}, a working model of situation characteristics showing that human behavior varies with situation characteristics. We construct scenario axes from nine pre-defined scenario labels. At the end, five scenario axes are constructed, corresponding to five of the eight DIAMONDS - Duty, Intellect, Adversity, Negativity, and Sociality \citep{rauthmann2014diamonds}. Evaluations across fixed and random temperature policies, nucleus sampling \citep{holtzman2020curious}, and EDT \citep{zhang2024edt} baselines indicate Scenario-Actor's competitive performance on interactive social scenarios, open-ended instruction following, and multi-turn dialogue compared to most decoding strategies. 

\section{Background}

An autoregressive language model generates a response one token at a time. In each next-token generation, a trained neural network first produces a set of scores, which we call logits, from the prompt $\mathcal{P}$ and the generated tokens over the vocabulary $\mathcal{V}$ that represents all possible next tokens. A $\operatorname{Softmax}$ function then transforms these logits to a distribution that gives how likely a token is to be the next. Let $y_t$ and $y_{<t}$ be the $t$-th token to be generated and already generated, and $z_{t,v}$ be the $t$-th logits of the token $v$ in the vocabulary. Concretely, we have
\[
\Pr(y_t=v \mid P,y_{<t})
=
\frac{\exp(z_{t,v})}
{\sum_{u\in\mathcal{V}}\exp(z_{t,u})}
\]
Now, notice that if we divide each logit $z_{t,v}$ with a positive scalar $T$, the relative contribution of each numerator to the denominator will change with $T$. When $T>1$, each $\exp(z_{t,v})$ becomes less dominant over $\sum_{u\in\mathcal{V}}\exp(z_{t,u})$, making the probabilities more evenly distributed. When $0<T<1$, each $\exp(z_{t,v})$ becomes more dominant. This scalar therefore controls the sharpness of the next-token distribution, which is commonly referred to as the temperature. Formally, for $T>0$,
\[
\Pr(y_t=v \mid P,y_{<t})
=
\frac{\exp(\frac{z_{t,v}}{T})}
{\sum_{u\in\mathcal{V}}\exp(\frac{z_{t,u}}{T})}
\]
With all these, the last thing we need to do is to select a token through a process called sampling. Recent work center around language model's temperature and sampling strategies, which we discuss in the following paragraphs.

\paragraph{Fixed decoding hyperparameters.} Early work and much of engineering practice predominantly relied on manually chosen global decoding strategies for task-specific models, such as beam search, top-k, and fixed sampling temperatures. \citet{holtzman2020curious} showed that in open-ended text generation, likelihood-maximizing decoding methods often lead to bland, repetitive, and degenerate outputs, and proposed nucleus sampling (top-$p$) as a more effective alternative. \citet{zhang2021trading} examined the trade-off between quality and diversity in different decoding strategies. Their findings indicate that all decoding methods perform similarly when diversity is a priority, but nucleus sampling \citep{holtzman2020curious} consistently outperforms all other evaluated algorithms when quality is viewed as more important. \citet{wiher2022decoding} revealed that the effectiveness of decoding strategies is highly task-dependent. Later in LLM era, \citet{shi2024thorough} made a more systematic horizontal comparison among decoding methods and revealed that decoding method performance strongly depends on task, model size, alignment, and quantization.

\begin{figure*}[t]
    \begin{minipage}{0.5\textwidth}
        \centering
        \includegraphics[width=\columnwidth]{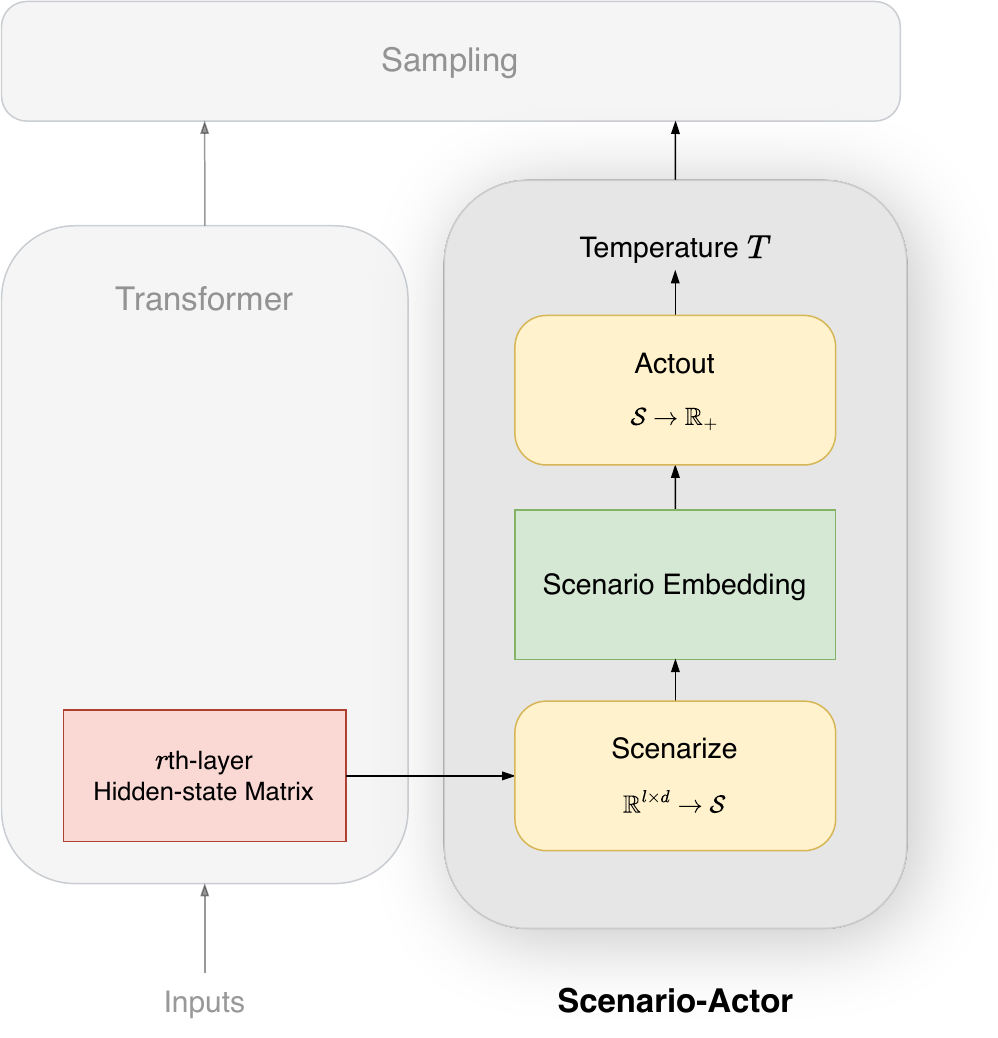}
        \caption{The Scenario-Actor - framework overview.}
        \label{fig:scenario-actor}
    \end{minipage}
    \begin{minipage}{0.5\textwidth}
        \centering
        \includegraphics[width=\columnwidth]{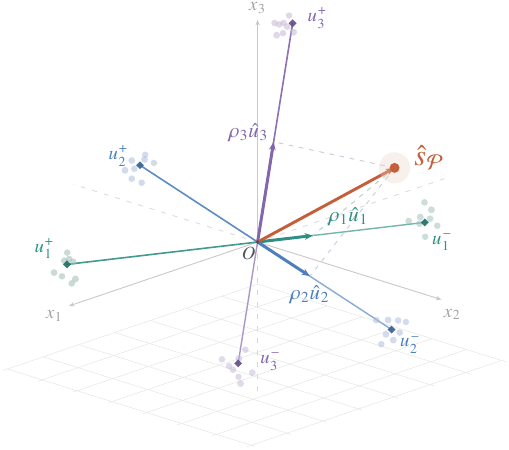}
        \caption{Visualization of scenario space and $\operatorname{Actout}$.}
        \label{fig:geometry-visualization}
    \end{minipage}
\end{figure*}

\paragraph{Token-level dynamic temperature.} Fixed decoding hyperparameters are not always optimal at every decoding step. 
An early token-level architecture begins with \citet{chang2023kl}, who proposed to relax the fixed-temperature assumption and adjust temperature dynamically according to its relevance to the source through KL-divergence. \citet{zhu2024hot} then introduced AdapT sampling for code generation. \citet{zhang2024edt} further put forth EDT, which adjusts temperature directly from token-level entropy to better balance quality and diversity. 

\paragraph{Context and example-level temperature.} Another line of work concerns selecting sampling temperature from context. \citet{wang2020contextual} proposed contextual temperature, which learns a context-dependent temperature vector over the output vocabulary. \citet{li2024dynamic} proposed DDS for open-domain dialogue system. DDS copes with two situations - chit-chat and factual QA - and maps context-dependent diversity score to a single decoding temperature. \citet{dhuliawala2024adaptive} introduced adaptive decoding, which allows temperature to be predicted either once per example or dynamically at each decoding step.

\section{The Scenario-Actor}

\label{sec:scenario-actor}
Most modern LLMs are built using the celebrated Transformer architecture. In our framework, we assume that our prompt context is represented by a Transformer model and take the $r$-th layer hidden-state matrix as input. The Scenario-Actor has two fundamental parts: $\operatorname{Scenarize}$ and $\operatorname{Actout}$.\footnote{The module names \textit{Scenarize} and \textit{Actout} reflect our view that a prompt induces a latent scenario, and the model acts out, or decodes, from that scenario accordingly.} $\operatorname{Scenarize}$ takes the hidden-state matrix at the $r$-th Transformer layer as its input and outputs the \textbf{scenario embedding}; $\operatorname{Actout}$ takes a scenario embedding and maps it to a sampling temperature $T$.


In the remainder of this section, we discuss these two modules in depth.

\begin{figure*}[htbp]
    \hspace*{0.026\columnwidth}
    \includegraphics[width=0.99\textwidth,page=1]{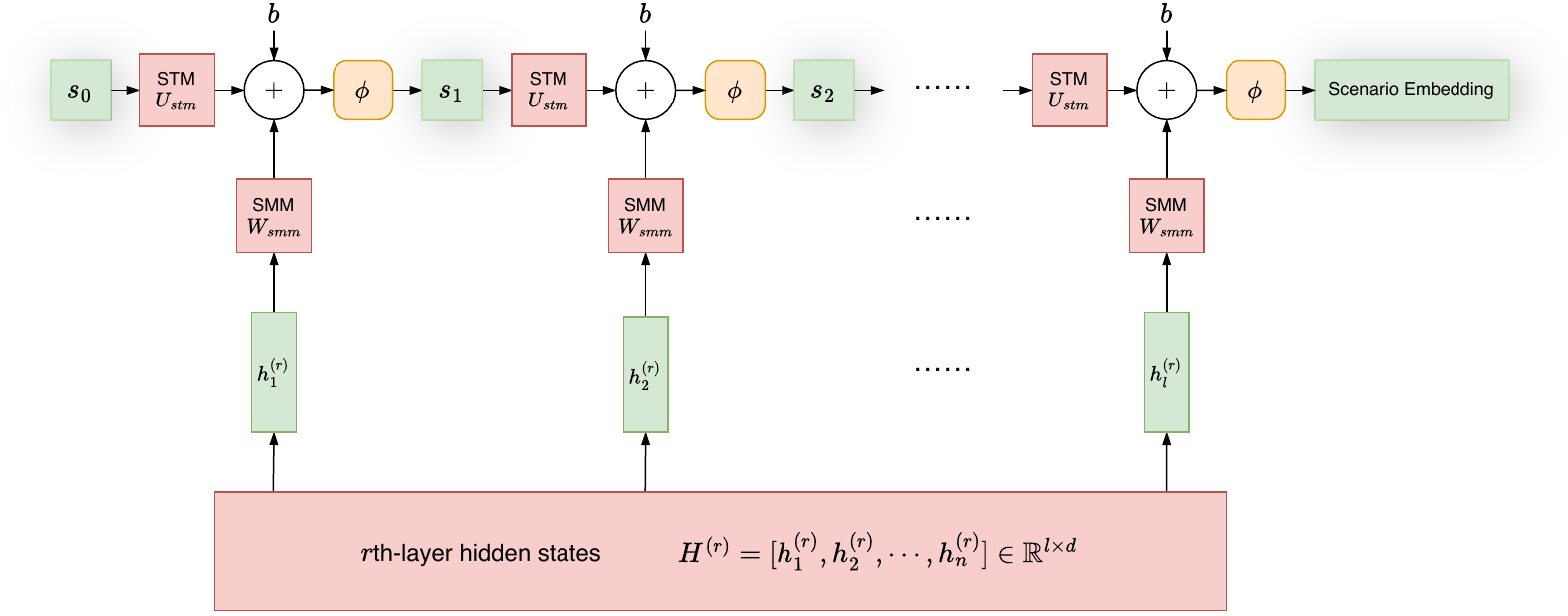}
    \caption{The $\operatorname{Scenarize}$ module.}
    \label{fig:scenarize}
\end{figure*}

\subsection{Scenarize}
Before introducing our first module, let's establish some basic mathematical setup. For a prompt $\mathcal{P}=\{p_1,\dots,p_l\}$ with $l$ tokens, let $H_{\mathcal{P}}^{(r)} = [h_{p_1}^{(r)},\dots,h_{p_l}^{(r)}]^T \in \mathbb{R}^{l \times d}$ denote the hidden-state matrix at the $r$-th Transformer layer, and $s_{\mathcal{P}} \in \mathcal{S} \subseteq \mathbb{R}^m$ denote the learned $m$-dimensional scenario embedding of prompt $\mathcal{P}$. The set $S \coloneqq \left\{ s_{\mathcal{P}} \;\middle|\; s_{\mathcal{P}} = \operatorname{Scenarize}(H_{\mathcal{P}}^{(r)}) \in \mathbb{R}^m \right\}$ be the set of all learned scenario embeddings. We define the scenario space as
\[
\mathcal{S} \coloneqq \operatorname{span}(S) \subseteq \mathbb{R}^m.
\]
Given that the language model interaction is text-only, the underlying prompt scenario is often under-determined with shorter linguistic sequences. As more tokens are observed, the scenario is generally expected to be more identifiable since longer prompts typically provide additional disambiguating context. As such, we model $\operatorname{Scenarize}(H^{(r)}_\mathcal{P})$ as a recurrent neural network (RNN) that updates the inferred scenario token by token. The final state at $p_l$ therefore represents the most informed scenario embedding which we will consider. Overall, we compute scenario embedding as
\begin{equation}
\label{eq:scenarize}
    \begin{aligned}
        s_t &= \operatorname{Scenarize}(h_{p_t}^{(r)}) \\ 
            &= \phi(U_{\text{stm}}s_{t-1}+W_{\text{smm}}h_{p_t}^{(r)} + b)
\end{aligned}
\end{equation}

Here, $\phi(\cdot)$ is the activation function; $U_{\text{stm}} \in \mathbb{R}^{m \times m}$ is the scenario transition matrix (STM); $W_{\text{smm}} \in \mathbb{R}^{m \times d}$ is the scenario mapping matrix (SMM); and $b \in \mathbb{R}^m$ is the bias term. Each scenario vector is computed as the learned affine combination of the previous token estimate and the current inference from context $h_{p_t}^{(r)} \in H_{\mathcal{P}}^{(r)}$, as depicted in Figure~\ref{fig:scenarize}.

Both STM and SMM are expected to be trained on prompt triplets $(\mathcal{P}, \mathcal{P}^{+}, \mathcal{P}^{-})$, where $\mathcal{P}$ denotes the prompt with base scenario, $\mathcal{P}^{+}$ the prompt with similar scenario, and $\mathcal{P}^{-}$ the prompt with opposite scenario. Given such a triplet, we define the loss function as
\begin{equation}
\label{eq:triplet_loss}
    \mathcal{L} = \max(0, d(s_\mathcal{P}, s_{\mathcal{P}^+})-d(s_{\mathcal{P}}, s_{\mathcal{P}^-}) + \beta) 
\end{equation}
which compares the distance from the base prompt scenario vector to the similar and opposite scenario vector with a margin $\beta$. In this work, we employ the squared Euclidean distance $d(s_\mathcal{P}, s_{\mathcal{P}^+})=\lVert s_{\mathcal{P}} - s_{\mathcal{P}^{+}} \rVert^2$ and $d(s_\mathcal{P}, s_{\mathcal{P}^-})=\lVert s_{\mathcal{P}} - s_{\mathcal{P}^{-}} \rVert^2$.

\subsection{Actout}

$\operatorname{Actout}$ then maps a scenario embedding to a temperature. Drawing on \textit{SemAxis} \citep{an2018semaxis}, we define scenario axes from two non-zero opposite scenario vectors $s^+$ and $s^-$. Each scenario axis $u_i$ can be defined as $u_i = s_i^+ - s_i^- \in \mathbb{R}^m$. We name the scenario vectors on all scenario axes as \textbf{subscenario vectors}, each representing an elementary component of a more complex scenario. For all scenario axes, two typical temperatures $\tau_i^+$ and $\tau_i^-$ on both the positive and negative directions are applied. In general, $u_1, \dots, u_k$ have one-to-one correspondence with $(\tau_1^+, \tau_1^-), \dots, (\tau_k^+, \tau_k^-) \in \mathbb{R}_{\geq 0} \times \mathbb{R}_{\geq 0}$. 

Let $\rho_i = \hat{s}_{\mathcal{P}}^\top \hat{u}_i$ be the signed cosine similarity between the normalized scenario embedding and the axis. For $\sum_{i=1}^{k} |\rho_i| > 0$, the temperature $T$ is then defined as
\begin{equation}
T =
\frac{
\sum_{i=1}^{k}
|\rho_i|
\left(
\frac{1+\rho_i}{2}\tau_i^+
+
\frac{1-\rho_i}{2}\tau_i^-
\right)
}{
\sum_{i=1}^{k}
|\rho_i|
}
\label{eq:actout}
\end{equation}
Temperature is defined as a normalized weighted average over $k$ scenario axes. For each axis $\hat{u}_i$, the cosine similarity $\rho_i$ specifies where the scenario embedding lies along that axis. Endpoint temperatures $\tau_i^+$ and $\tau_i^-$ are linearly interpolated based on $\rho_i$, so that the estimate moves towards $\tau_i^+$ if $\rho_i$ is positive and $\tau_i^-$ if negative. The denominator normalizes the weighted sum, eventually yielding a temperature inferred from the prompt's position in scenario space.


\section{Experiment}

We evaluate our proposed Scenario-Actor framework on benchmarks covering interactive social scenarios, open-ended instruction following and multi-turn question answering to test its performance on context-rich and socially-situated interactions. We also use GSM8K \citep{cobbe2021gsm8k} as a task-only stress test to study the boundary of scenario-based decoding.

\subsection{Experimental Setup}

\subsubsection{Backbone model} Despite the outstanding performance of mainstream LLMs such as ChatGPT and Gemini, we do not have access to the hidden states with API calls. Therefore, we use the open-weight Llama-3.1-8B-Instruct \citep{grattafiori2024llama3} as our backbone model. The model is used for both hidden-state extraction and generation. 

\subsubsection{Data} 
We collect 37{,}357 conversational prompts from human-human and human-LLM interaction corpora, including WildChat~\citep{zhao2024wildchat} and DailyDialog~\citep{li2017dailydialog}, together with targeted supplementary datasets for under-represented scenario types. Each prompt is annotated by GPT-4o~\citep{openai2024gpt4o} with one of nine scenario labels. A confidence-based filtering is applied to construct a high-confidence supervision set, resulting in 34,427 training prompts for triplet construction and 2,930 validation prompts.

\subsubsection{Baselines} We compare Scenario-Actor against fixed-temperature decoding at $T \in \{0.5, 0.7, 1.0, 1.5\}$, greedy decoding, random temperature selection, nucleus sampling~\citep{holtzman2020curious} with $p = 0.95$ and $T=1.0$, and entropy-driven temperature decoding (EDT) \citep{zhang2024edt}. EDT is included as a training-free and directly reproducible adaptive baseline. DDS and LPO \citep{li2024dynamic, dhuliawala2024adaptive} both, in contrast, require training additional components. LPO trains its decoder module via preference optimization over sampled outputs, and DDS trains a diversity regressor designed for a two-way chit-chat vs. QA distinction, with, to our knowledge, no public implementation. Reproducing either method involves substantial heuristic design choices of our own, making the comparison hard to attribute in either direction.

\subsubsection{Implementation details}
\paragraph{Hidden states} Llama-3.1-8B has 32 Transformer layers. For all experiments, we use the hidden states at the 16th Transformer layer. We found that scenario geometry collapses with deeper Transformer layer, which we discuss in Section \ref{sec:ablation}.

\paragraph{Scenarize} $\operatorname{Scenarize}$ maps the above hidden states into a 32-dimensional scenario embedding. The implementation follows Equation~\ref{eq:scenarize}, where $r=16$ is the Transformer layer, $\phi(\cdot) = \operatorname{tanh}(\cdot)$ is the activation function. $\operatorname{Scenarize}$ is trained according to Equation~\ref{eq:triplet_loss} with squared Euclidean triplet loss with margin $\beta=0.3$. AdamW with learning rate 3.0e-4, weight decay 1.0e-4, gradient clipping at 1.0, and early stopping based on validation recall@1 is used as optimizer.

\begin{table*}[t]
\centering
\setlength{\tabcolsep}{12pt}
\begin{tabular}{lccc}
\toprule
\textbf{Baseline}
& \textbf{AgentSense}
& \textbf{AlpacaEval}
& \textbf{MT-Bench} \\
\midrule
\textit{Metric}
& Goal completion
& LC-WR (\%)
& GPT-4 score \\
\midrule
Scenario-Actor
& \textbf{0.8015}
& \textbf{30.03}
& \textbf{7.91} \\
Greedy
& 0.7582
& 22.87
& 7.79 \\
Fixed $T=0.5$
& 0.7540
& 23.63
& 7.79 \\
Fixed $T=0.7$
& 0.7801
& 25.68
& 7.83 \\
Fixed $T=1.0$
& 0.7692
& 23.42
& 7.61 \\
Fixed $T=1.5$
& 0.6989
& 8.60
& 6.76 \\
EDT
& 0.7671
& 25.58
& 6.74 \\
Nucleus $(p=0.95)$
& 0.7557
& 23.59
& 7.63 \\
Random-$T$
& 0.7699
& 20.44
& 7.71 \\
\bottomrule
\end{tabular}
\caption{Benchmark results comparing Scenario-Actor with fixed-temperature and adaptive decoding baselines. Best results are \textbf{highlighted}.}
\label{tab:baseline_comparison}
\end{table*}

\paragraph{Actout} We operationalize five DIAMONDS dimensions with nine scenario labels. We compute the centroid of each scenario label as a subscenario vector and use these vectors to define each scenario axes. GPT-4o \citep{openai2024gpt4o} is used as an LLM annotator for the training set. The nine scenario labels are rigorous, brainstorming, formal, playful, adversarial, institutional, casual\_conversation, emotional\_support\_pos, and emotional\_support\_neg.

\begin{table}[H]
\centering
\begin{tabular}{lc}
\toprule
\textbf{Scenario Label} & $\boldsymbol{\tau_i^+ / \tau_i^-}$ \\
\midrule
rigorous & 0.30 \\
institutional & 0.50 \\
formal & 0.55 \\
emotional\_support\_neg & 0.65 \\
adversarial & 0.75 \\
emotional\_support\_pos & 1.00 \\
casual\_conversation & 1.10 \\
brainstorming & 1.40 \\
playful & 1.50 \\
\bottomrule
\end{tabular}
\caption{Scenario labels and their endpoint temperatures.}
\label{tab:scenario_temperature}
\end{table}

Five scenario axes - institutional vs. casual\_conversation, rigorous vs. brainstorming, adversarial vs. non-adversarial; emotional\_support\_pos vs. emotional\_support\_neg, and formal vs. playful were constructed from combinations of nine subscenario vectors. Non-adversarial is the average over all subscenario vectors that are not adversarial. Detailed description of each scenario can be found in Appendix~\ref{app:scenario-descriptions}. Such five scenario axes match five of the DIAMONDS: Duty, Intellect, Adversity, Negativity, and Sociality \citep{rauthmann2014diamonds}.

Each of the five axes is assigned two endpoint temperatures $(\tau_i^+, \tau_i^-)$, which are external hyperparameters of $\operatorname{Actout}$, as shown in Table~\ref{tab:scenario_temperature}. $\operatorname{Actout}$ then maps a scenario embedding generated from $\operatorname{Scenarize}$ to a temperature within the range $[0.30, 1.50]$ corresponding to Equation~\ref{eq:actout}. In our experiment, we fixed $(\tau_i^+, \tau_i^-)$ for all benchmarks and comparisons.

\subsection{Main Result}
We evaluate Scenario-Actor across three benchmarks - AgentSense \citep{mou2025agentsense}, AlpacaEval \citep{dubois2024alpacaeval}, and MT-Bench \citep{zheng2023mtbench} - covering situational goal-completion, open-ended instruction following, and multi-turn interactions with respect to all baselines. Goal completion score, Length-Controlled Win Rate (LC-WR), and GPT-4 score were metrics for each of the three main evaluations. The performance is shown in Table~\ref{tab:baseline_comparison}. Scenario-Actor significantly outperforms most decoding methods on AgentSense and all baselines on AlpacaEval, while remaining competitive on MT-Bench in our experiment.

Among available benchmarks, AgentSense directly evaluates the setting targeted by Scenario-Actor: interactive dialogue in which the model must adapt to complete different goals as the social context develops. Scenario-Actor achieves the highest goal completion score of 0.8015, outperforming greedy decoding, every fixed-temperature setting, EDT, nucleus sampling, and Random-$T$. Scenario-Actor attains significant improvements over all baselines on the range from 3.16 to 10.26 percentage points except for fixed $T=0.7$ at 0.7801, for which Scenario-Actor's 2.14 percentage points advantage is not statistically significant.

On AlpacaEval, Scenario-Actor achieves the strongest result among all tested methods with 30.03\% Length-Controlled Win Rate (LC-WR), outperforming the best fixed-temperature baseline by 4.35 percentage points, EDT by 4.45 percentage points, and exceeding nucleus sampling, greedy decoding, Random-$T$, and class-conditioned $T$ with statistical significance. On MT-Bench, our framework scores 7.91, which is significantly above fixed $T=1.0$, fixed $T=1.5$, Random-$T$, nucleus sampling, EDT, and class-conditioned $T$, and is potentially competitive among greedy decoding, fixed $T=0.5$, and fixed $T=0.7$. Detailed statistical report can be found in Appendix~\ref{app:significance}.

\subsection{Analysis}

\begin{figure*}[t]
    \begin{minipage}{0.5\textwidth}
        \centering
        \includegraphics[width=\linewidth,page=1]{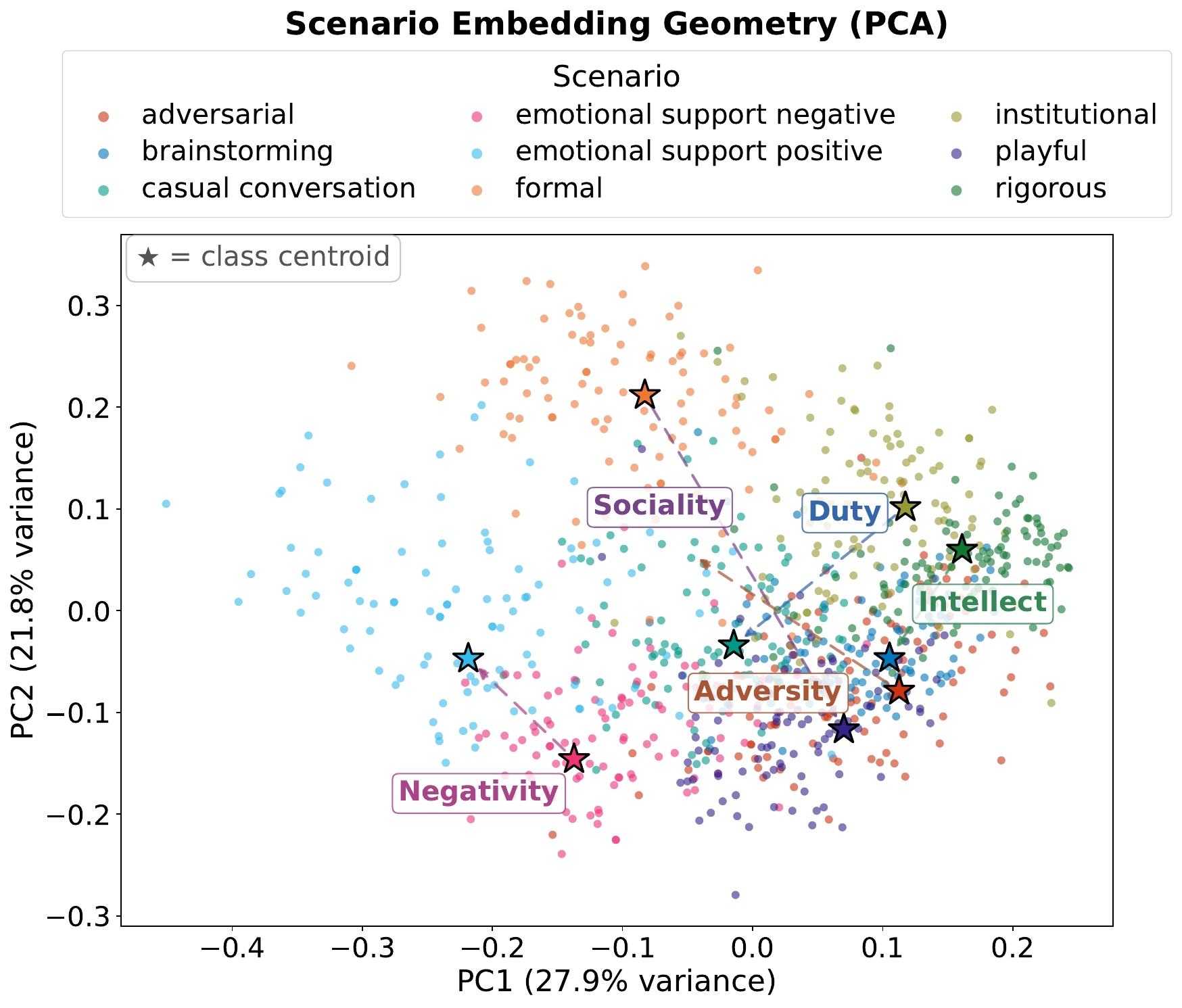}
    \end{minipage}
    \begin{minipage}{0.5\textwidth}
        \centering
        \includegraphics[width=\linewidth,page=1]{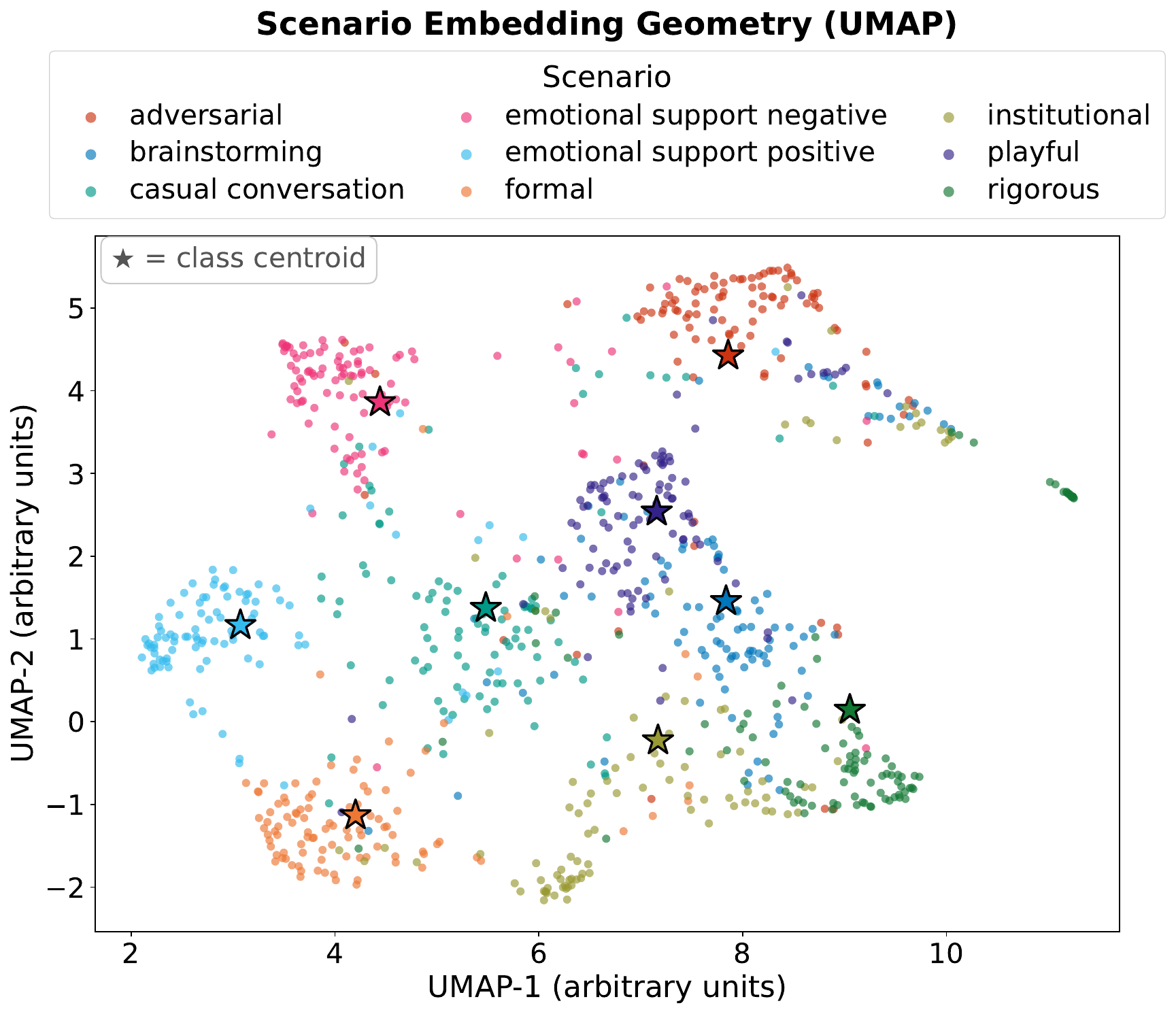}
    \end{minipage}
    \caption{PCA (left) and UMAP (right) of scenario embedding geometry.}
    \label{fig:scenario-embedding-pca-umap}
\end{figure*}

\paragraph{Agreements}

To justify the use of LLM for scenario annotation and results judging, we compute both the LLM-human and human-human agreement on the models we used. A gold set on scenario annotation consisting of 306 prompts and a gold set on evaluation consisting of 50 AlpacaEval examples and 50 MT-Bench examples were manually annotated by us. 

\begin{table}[h]
\centering
\setlength{\tabcolsep}{4pt}
\begin{tabular}{lcc}
\hline
Validation & LLM-human & Human-human \\
\hline
Annotation $\kappa$ & 0.82 / 0.85 & 0.79 \\
AE Agreement & 86.00 / 86.00 & 80.00 \\
AE judge $\kappa$ & 0.72 / 0.72 & 0.60 \\
MT Pearson $r$ & 0.92 / 0.86 & 0.87 \\
MT Spearman $\rho$ & 0.84 / 0.70 & 0.79 \\
AS Agreement & 80.00 / 72.00 & 76.00 \\
AS Cohen's $\kappa$ & 0.48 / 0.27 & 0.40 \\
\hline
\end{tabular}
\caption{Agreement between LLM-based evaluators and human annotators. Values are rounded to two decimal places. Detailed validation metrics are reported in Appendix~\ref{app:agreements}.}
\label{tab:agreement-validation}
\end{table}

The results show that the GPT-4o annotator agrees with each human annotator at $\kappa = 0.82$ and $0.85$. The GPT-4-turbo \citep{openai2023gpt4} AlpacaEval judge reaches $\kappa \approx 0.72$ with each annotator; for MT-Bench scoring, it correlates with human raters at Pearson $r = 0.92$ and $0.86$; for AgentSense, it achieves agreement rates of 80\% and 72\% with the two human judges, comparable to the human-human agreement of 76\%, and with Cohen’s $\kappa$ of 0.48 and 0.27, respectively.

\paragraph{Geometry} To validate if $\operatorname{Scenarize}$ learns meaningful scenario geometry, we visualize the learned scenario space using both PCA and UMAP. PCA shows the global variance structure of 900 scenario embeddings, with 100 examples from each scenario label. The first two principal components explain 27.9\% and 21.8\% of the variance respectively. As shown in Figure~\ref{fig:scenario-embedding-pca-umap}, nine centroids obtained from nine scenario labels occupy different regions in scenario space. PCA view also shows the five proposed scenario axes used in our experiment - Duty, Intellect, Negativity, Sociality, and Adversity. The patterns therefore reflect that $\operatorname{Scenarize}$ learns a structured scenario space.

UMAP further confirms local structure of the learned space. The projection achieves a trustworthiness score of 0.96 using cosine distance. The embedding exhibits meaningful relationships among the nine scenario types. Semantically related scenarios also cluster near one another; for example, institutional prompts lie closer to rigorous prompts than to casual conversation.

\paragraph{GSM8K diagnosis} To study the boundary of Scenario-Actor, we tested its performance on GSM8K against all other baselines. Scenario-Actor significantly underperforms greedy decoding, fixed $T=0.5$, fixed $T=0.7$, EDT, and insignificantly lower than Random-$T$ baseline by 0.53.
\begin{table}[H]
\centering
\setlength{\tabcolsep}{6pt}
\begin{tabular}{lc}
\toprule
\textbf{Baseline}
& \textbf{GSM8K} \\
\midrule
\textit{Metric}
& Accuracy (\%) \\
\midrule
Scenario-Actor
& 65.28 \\
Greedy
& \textbf{77.18} \\
Fixed $T=0.5$
& 69.52 \\
Fixed $T=0.7$
& 69.07 \\
Fixed $T=1.0$
& 62.40 \\
Fixed $T=1.5$
& 45.19 \\
EDT
& 69.14 \\
Nucleus $(p=0.95)$
& 61.33 \\
Random-$T$
& 65.81 \\
\bottomrule
\end{tabular}
\caption{GSM8K results comparing Scenario-Actor with fixed-temperature and adaptive decoding baselines. The best result is \textbf{highlighted}.}
\label{tab:gsm8k_comparison}
\end{table}

Our GSM8K temperature is distributed within the range of $[0.8162, 0.8782]$. In theory, GSM8K would exhibit better performance of the model when the sampling temperature is lower. In our experiment, Scenario-Actor selects a GSM8K temperature less than ideal. Nearest-centroid analysis shows that the distribution is concentrated in adversarial, institutional, and brainstorming scenarios, which together account for 77.9\% of GSM8K prompts. Only 5.0\% are mapped to the rigorous scenario.
\begin{table}[H]
\centering
\begin{tabular}{lcc}
\toprule
\textbf{Scenario}
& \textbf{Count}
& \textbf{Percentage} \\
\midrule
adversarial
& 350
& 26.5\% \\
institutional
& 346
& 26.2\% \\
brainstorming
& 332
& 25.2\% \\
playful
& 79
& 6.0\% \\
casual\_conversation
& 67
& 5.1\% \\
rigorous
& 66
& 5.0\% \\
formal
& 40
& 3.0\% \\
emotional\_support\_neg
& 23
& 1.7\% \\
emotional\_support\_pos
& 16
& 1.2\% \\
\bottomrule
\end{tabular}
\caption{GSM8K scenario count}
\label{tab:gsm8k-scenario-count}
\end{table}
\FloatBarrier
\noindent All nine centroids look nearly identical to GSM8K prompts in cosine space:
\begin{table}[H]
\centering
\setlength{\tabcolsep}{9pt}
\begin{tabular}{lc}
\toprule
\textbf{Scenario Centroid}
& \textbf{Cosine Similarity} \\
\midrule
adversarial
& 0.9765 \\
institutional
& 0.9710 \\
rigorous
& 0.9684 \\
brainstorming
& 0.9510 \\
playful
& 0.9499 \\
casual\_conversation
& 0.9463 \\
emotional\_support\_neg
& 0.9191 \\
formal
& 0.9098 \\
emotional\_support\_pos
& 0.9043 \\
\bottomrule
\end{tabular}
\caption{Mean cosine similarity between GSM8K and nine centroids, with gap of 0.0247 between the nearest and the second-nearest centroid per prompt}
\label{tab:gsm8k-cosine-similarity}
\end{table}
The results reflect near-equidistance of GSM8K to all scenario centroids: GSM8K prompts occupy a high-similarity region of the learned space. The current geometry does not reliably map task-only mathematical prompts to a low-temperature rigorous regime, and Scenario-Actor fails to infer the “scenario” of GSM8K prompts. The result illustrates the scope and boundary of Scenario-Actor to be a decoding framework for prompts whose communicative scenario can be inferred from context.

\paragraph{Temperature} We examine the temperatures selected by Scenario-Actor across all benchmarks. Scenario-Actor operates within the range of $[0.67, 1.16]$ across turns generated by the evaluated Llama-3.1-8B-Instruct, $[0.77, 1.01]$ on AlpacaEval, and $[0.76, 1.02]$ on MT-Bench. AgentSense temperatures also change with dialogue context, with the mean temperature increasing by approximately 0.16 across turn positions.

\paragraph{Context length} We compare outputs generated by Scenario-Actor under short context with only one prior turn versus long context with all prior turns on 100 DailyDialog multi-turn conversations. Long context Scenario-Actor achieves a win rate of 67.0\% over short context with a 95\% bootstrap confidence interval of [58.0\%, 76.0\%] in pairwise GPT-4-turbo evaluation. The finding supports our claim that scenario is generally expected to be more identifiable with more context, which motivates the RNN formulation of $\operatorname{Scenarize}$.

\paragraph{Efficiency}
We measure the latency, memory usage, and throughput of Scenario-Actor over 80 MT-Bench prompts on a single NVIDIA A10G 24 GB GPU (\texttt{ml.g5.2xlarge}). Latency includes hidden-state extraction, Scenarize, Actout, and response generation. As shown in Appendix~\ref{app:efficiency}, Scenario-Actor has similar end-to-end latency and throughput to the tested decoding baselines. Latency is within 2.5\% of greedy decoding, while the output and request throughput remain close to standard sampling methods. Peak allocated memory is also nearly unchanged.

\subsection{Ablation}
\label{sec:ablation}
\paragraph{Geometry contributes more than endpoint temperature lookup.} To test whether the gains come from the scenario geometry or from the manually-specified endpoint temperatures, we add a class-conditioned temperature baseline which uses the same $\operatorname{Scenarize}$ module but removes $\operatorname{Actout}$'s geometric weighting. Class-conditioned baseline treats the nearest centroid as the prompt's class label and assigns its sampling temperature from corresponding endpoint.

\begin{table}[H]
\centering
\setlength{\tabcolsep}{3pt}
\begin{tabular*}{\columnwidth}{@{}lcc@{}}
\toprule
& \multicolumn{2}{c}{\textbf{Baseline}} \\
\cmidrule(l){2-3}
\textbf{Benchmark}
& Scenario-Actor
& Class-conditioned $T$ \\
\midrule
\textbf{AgentSense}
& \textbf{0.8015}
& 0.7536 \\
\textbf{AlpacaEval}
& \textbf{30.03}
& 27.10 \\
\textbf{MT-Bench}
& \textbf{7.91}
& 7.64 \\
\bottomrule
\end{tabular*}
\caption{Comparison with the class-conditioned temperature baseline.
Metrics are goal completion score, LC-WR (\%), and GPT-4 score respectively. Best results are \textbf{highlighted}.}
\label{tab:class_conditioned_comparison}
\end{table}
\FloatBarrier

Class-conditioned temperature baseline achieves a goal completion score of 0.7536 on AgentSense, LC-WR of 27.10\% on AlpacaEval and a score of 7.64 on MT-Bench, which is lower than Scenario-Actor by 4.79 and 2.93 percentage points, and 0.27 points respectively. The improvements are statistically significant on all three benchmarks: AgentSense ($[+1.85,+7.88]$, $p=1.85 \times 10^{-3}$), AlpacaEval ($[+0.55,+5.32]$, $p=1.6 \times 10^{-2}$), and MT-Bench ($[+0.02,+0.52]$, $p=3.21 \times 10^{-2}$). Since both methods use the same endpoint values, these results suggest the effectiveness of Actout's geometric interpolation compared to class-level temperature lookup.

\paragraph{Layer 16 provides the best tradeoff for scenario probing.} To test which layer would be the optimal for scenario decoding, we evaluate several metrics across different Transformer layers. Four of which are \textbf{recall@1}, which computes how often the nearest retrieved embedding belongs to the same scenario; \textbf{effective rank}, measuring how many embedding dimensions are meaningfully used; \textbf{tanh saturation}, which calculates the percentage of embedding values close to $-1$ or $1$; and \textbf{axis correlation} that gives the maximum correlation between scenario axes.

We observe that layer 16 produces competitive scenario retrieval performance, with stronger scenario axis separation and lower representation saturation than deeper layers such as layer 30, as shown in Table~\ref{tab:layer-sweep} below. This observation aligns with recent work suggesting that structured semantic relation signals in LLMs often peak in intermediate layers rather than final generation-oriented layers \citep{diera2026semanticrelations}. Full layer-sweep diagnostics are reported in Appendix~\ref{app:layer-sweep}.
\begin{table}[H]
\centering
\setlength{\tabcolsep}{6pt}
\begin{tabular}{lccc}
\toprule
Metric & Layer 16 & Layer 20 & Layer 30 \\
\midrule
Recall@1 & 0.770 & \textbf{0.798} & 0.518 \\
Eff. Rank & \textbf{24.41} & 22.80 & 7.08 \\
tanh Sat. & \textbf{4.33\%} & 15.42\% & 92.24\% \\
Axis Corr. & \textbf{0.60} & 0.80 & 0.94 \\
\bottomrule
\end{tabular}
\caption{Layer-sweep diagnostics. Best results are \textbf{highlighted}}
\label{tab:layer-sweep}
\end{table}

\paragraph{Larger triplet margins induce geometry-performance tradeoffs.}
We compare the default margin $\beta=0.3$ against $\beta=1.0$ while holding the architecture, training data, and inference protocol fixed. As in Table~\ref{tab:margin_ablation}, increasing margin produces stronger centroid separation and improves GSM8K accuracy, but reduces performance on multi-turn and open-ended evaluation tasks. These results suggest that larger margins reshape the scenario geometry but does not uniformly improve generation quality.
\begin{table}[H]
\centering
\setlength{\tabcolsep}{4pt}
\begin{tabular}{@{}lcc@{}}
\toprule
\textbf{Metric}
& $\boldsymbol{\beta=0.3}$
& $\boldsymbol{\beta=1.0}$ \\
\midrule
\multicolumn{3}{@{}l}{\textit{Geometry}} \\
Mean centroid cosine
& 0.9496 & 0.7103 \\
Euclidean silhouette
& 0.1835 & 0.2668 \\
Nearest-centroid agreement
& 0.6940 & 0.8877 \\
\midrule
\multicolumn{3}{@{}l}{\textit{Downstream}} \\
GSM8K Accuracy (\%)
& 65.28 & 66.41 \\
AlpacaEval LC-WR (\%)
& 30.03 & 25.54 \\
MT-Bench
& 7.91 & 7.66 \\
GSM8K mean $T$
& 0.822 & 0.738 \\
\bottomrule
\end{tabular}
\caption{Effect of the triplet-loss margin on scenario geometry and downstream performance. Full table can be found in Appendix~\ref{app:beta-margin}.}
\label{tab:margin_ablation}
\end{table}

\section{Conclusion}
We propose Scenario-Actor, a sampling temperature selection framework that maps a hidden-state matrix to a structured scenario embedding and derives a sampling temperature from its geometry along interpretable scenario axes. Comparisons of Scenario-Actor with multiple baselines show strong goal-completion performance on AgentSense, significant improvements on AlpacaEval, and competitive results on MT-Bench. In addition, we see Scenario-Actor as a step toward socially adaptive language models.  As LLMs become embedded in human society through multimodal agents and embodied systems, their ability to recognize the situations they are in will become increasingly important. We hope this framework allows models to better understand social context, emotional nuance, and the expectations that shape human communication in the future.

\section*{Limitations}
\paragraph{LLM scenario is narrower than human scenario.} Language models take only textual sequence as their input. In actual human-human interaction, scenarios are dependent on cues other than pure linguistic context. Therefore, current language models are still limited in their ability to fully integrate into all aspects of human interactions. The inference of scenario solely from linguistic context is itself insufficient.

\paragraph{Compromise on task-only prompts.} Scenario-Actor is designed for settings with rich scenario cues, such as multi-turn conversations. But in our dataset, we see a great amount of them being straightforward and utilitarian. Many prompts ask the model to directly solve problems such as writing code, answering math or providing general information. Such a characteristic makes scenario inference almost impossible as task-oriented prompts often contain little evidence for a social setting. For these prompts, a standalone selector can route the input to a standard task-specific decoding policy, which is independent of our framework. This setup is also a compromise to current LLM usage as the language model itself is limited in its interaction modality. We expect this limitation to be alleviated in multimodal or embodied agent settings, where models can access richer interactional cues and become more deeply integrated into human social environments.

\paragraph{Model and taxonomy scope.} The current study establishes the results and claims merely on Llama-3.1-8B-Instruct with the nine-label, five-axis scenario taxonomy. In theory, however, our framework is expected to be applicable to other model families and scenario taxonomies, which, due to limited budget and computational power, require more evaluations.

\paragraph{Source balance.} A key limitation of the current training data is that scenario labels are not fully balanced from data source. In an ideal setting, each scenario category would be represented across multiple corpora so that $\operatorname{Scenarize}$ would be forced to learn merely the scenario structure rather than dataset style. In practice, however, no available dataset covers all of our scenario labels with balanced coverage. We therefore rely on multiple supplementary sources. Developing a more balanced scenario dataset remains an important direction for future work.

\section*{Acknowledgments}
We thank Professor Haifeng Xu at the University of Chicago and the anonymous reviewers for their valuable comments, as well as Professor Adam Meyers at New York University for his advice on publication. This work was conducted independently by the authors and received no institutional research funding by the time of submission.

\bibliography{all}

\clearpage
\appendix

\makeatletter
\setlength{\@dblfptop}{0pt}
\setlength{\@fptop}{0pt}
\makeatother

\begin{table*}[!t]
\begin{minipage}{\textwidth}
\section{Efficiency}
\label{app:efficiency}
\centering
\begin{tabular}{@{}lrrrrr@{}}
\toprule
\multirow{2}{*}{\textbf{Method}}
&
\multirow{2}{*}{\textbf{\shortstack{Latency\\(ms/episode)}}}
&
\multicolumn{2}{c}{\textbf{Peak memory (GiB)}}
&
\multicolumn{2}{c@{}}{\textbf{Throughput}} \\
\cmidrule(lr){3-4}
\cmidrule(l){5-6}
&
&
\textbf{Allocated}
&
\textbf{Reserved}
&
\textbf{\shortstack{Output\\(tokens/s)}}
&
\textbf{\shortstack{Request\\(episodes/s)}} \\
\midrule
\textbf{Scenario-Actor}
& 97,000 & 18.59 & 18.71 & 21.59 & 0.01031 \\
Greedy
& 87,410 & 18.36 & 18.48 & 21.84 & 0.01144 \\
Fixed $T=0.5$
& 88,110 & 18.38 & 18.50 & 21.57 & 0.01135 \\
Nucleus ($p=0.95$)
& 88,370 & 18.38 & 18.50 & 21.49 & 0.01132 \\
Random-$T$
& 87,340 & 18.38 & 18.50 & 21.59 & 0.01145 \\
EDT
& 87,650 & 18.38 & 18.55 & 21.64 & 0.01141 \\
\bottomrule
\end{tabular}
\caption{End-to-end efficiency over 245 AgentSense templates in 15-turn closed-loop interactions.}

\begin{tabular}{@{}lrrrrr@{}}
\toprule
\multirow{2}{*}{\textbf{Method}}
&
\multirow{2}{*}{\textbf{\shortstack{Latency\\(ms/prompt)}}}
&
\multicolumn{2}{c}{\textbf{Peak memory (GiB)}}
&
\multicolumn{2}{c@{}}{\textbf{Throughput}} \\
\cmidrule(lr){3-4}
\cmidrule(l){5-6}
&
&
\textbf{Allocated}
&
\textbf{Reserved}
&
\textbf{\shortstack{Output\\(tokens/s)}}
&
\textbf{\shortstack{Request\\(prompts/s)}} \\
\midrule
\textbf{Scenario-Actor}
& 17,109 & 16.31 & 16.68 & 25.35 & 0.0584 \\
Greedy
& 17,181 & 16.52 & 16.86 & 25.87 & 0.0582 \\
Fixed $T=0.5$
& 17,150 & 16.52 & 16.86 & 25.43 & 0.0583 \\
Nucleus ($p=0.95$)
& 17,406 & 16.52 & 16.86 & 25.42 & 0.0575 \\
Random-$T$
& 17,284 & 16.52 & 16.86 & 25.42 & 0.0579 \\
EDT
& 17,336 & 16.52 & 16.86 & 25.59 & 0.0577 \\
\bottomrule
\end{tabular}
\caption{End-to-end efficiency over 805 AlpacaEval prompts.}
\begin{tabular}{@{}lrrrrr@{}}
\toprule
\multirow{2}{*}{\textbf{Method}}
&
\multirow{2}{*}{\textbf{\shortstack{Latency\\(ms/prompt)}}}
&
\multicolumn{2}{c}{\textbf{Peak memory (GiB)}}
&
\multicolumn{2}{c@{}}{\textbf{Throughput}} \\
\cmidrule(lr){3-4}
\cmidrule(l){5-6}
&
&
\textbf{Allocated}
&
\textbf{Reserved}
&
\textbf{\shortstack{Output\\(tokens/s)}}
&
\textbf{\shortstack{Request\\(prompts/s)}} \\
\midrule
\textbf{Scenario-Actor}
& 16,917 & 15.22 & 15.54 & 25.26 & 0.0591 \\
Greedy
& 16,513 & 15.19 & 15.26 & 25.75 & 0.0606 \\
Fixed $T=0.5$
& 17,128 & 15.19 & 15.54 & 25.24 & 0.0584 \\
Nucleus ($p=0.95$)
& 16,845 & 15.19 & 15.26 & 25.29 & 0.0594 \\
Random-$T$
& 16,267 & 15.19 & 15.26 & 25.28 & 0.0615 \\
EDT
& 16,957 & 15.24 & 21.99 & 25.41 & 0.0590 \\
\bottomrule
\end{tabular}
\caption{End-to-end efficiency over 80 MT-Bench prompts.}
\label{tab:efficiency}

\end{minipage}
\end{table*}
\FloatBarrier

\clearpage
\begin{table*}[!t]
\begin{minipage}{\textwidth}
\section{Statistical Significance Testing}
\label{app:significance}
\centering
\setlength{\tabcolsep}{3pt}
\begin{tabular}{@{}llccc@{}}
\toprule
\multirow{2}{*}{\textbf{Benchmark}}
&
\multirow{2}{*}{\textbf{Comparison}}
&
\multicolumn{3}{c@{}}{\textbf{Statistical significance}} \\
\cmidrule(l){3-5}
&
&
\textbf{Difference}
&
\textbf{95\% CI}
&
\textbf{$p$-value} \\
\midrule
\multirow{9}{*}{\textbf{\shortstack[l]{AgentSense\\($\Delta$ Goal\\completion)}}}
& Scenario-Actor vs Greedy
& $+4.33$ pp.
& $[+1.55, +7.15]$
& $2.44 \times 10^{-3}$ \\
& Scenario-Actor vs Fixed $T=0.5$
& $+4.75$ pp.
& $[+1.82, +7.76]$
& $1.72 \times 10^{-3}$ \\
& Scenario-Actor vs Fixed $T=0.7$
& $+2.14$ pp.
& $[-0.54, +4.89]$
& $1.22 \times 10^{-1}$ \\
& Scenario-Actor vs Fixed $T=1.0$
& $+3.23$ pp.
& $[+0.41, +6.11]$
& $2.63 \times 10^{-2}$ \\
& Scenario-Actor vs Fixed $T=1.5$
& $+10.26$ pp.
& $[+6.97, +13.53]$
& $8.73 \times 10^{-10}$ \\
& Scenario-Actor vs EDT
& $+3.44$ pp.
& $[+0.50, +6.47]$
& $2.39 \times 10^{-2}$ \\
& Scenario-Actor vs Nucleus
& $+4.58$ pp.
& $[+1.41, +7.70]$
& $4.31 \times 10^{-3}$ \\
& Scenario-Actor vs Random-$T$
& $+3.16$ pp.
& $[+0.40, +5.97]$
& $2.62 \times 10^{-2}$ \\
& Scenario-Actor vs Class-conditioned $T$
& $+4.79$ pp.
& $[+1.85, +7.88]$
& $1.85 \times 10^{-3}$ \\
\midrule
\multirow{9}{*}{\textbf{\shortstack[l]{AlpacaEval\\($\Delta$ LC-WR)}}}
& Scenario-Actor vs Greedy
& $+7.16$
& $[+3.14,+11.18]$
& $5 \times 10^{-4}$ \\
& Scenario-Actor vs Fixed $T=0.5$
& $+6.40$
& $[+2.48,+10.32]$
& $1.4 \times 10^{-3}$ \\
& Scenario-Actor vs Fixed $T=0.7$
& $+4.35$
& $[+0.43,+8.27]$
& $2.96 \times 10^{-2}$ \\
& Scenario-Actor vs Fixed $T=1.0$
& $+6.61$
& $[+2.89,+10.33]$
& $5 \times 10^{-4}$ \\
& Scenario-Actor vs Fixed $T=1.5$
& $+21.43$
& $[+16.63,+26.23]$
& $2.2 \times 10^{-18}$ \\
& Scenario-Actor vs EDT
& $+4.45$
& $[+0.43,+8.47]$
& $3 \times 10^{-2}$ \\
& Scenario-Actor vs Nucleus
& $+6.44$
& $[+2.62,+10.26]$
& $1 \times 10^{-3}$ \\
& Scenario-Actor vs Random-$T$
& $+9.59$
& $[+5.28,+13.90]$
& $1.3 \times 10^{-5}$ \\
& Scenario-Actor vs Class-conditioned $T$
& $+2.93$
& $[+0.55,+5.32]$
& $1.6 \times 10^{-2}$ \\
\midrule
\multirow{9}{*}{\textbf{\shortstack[l]{MT-Bench\\($\Delta$ Score)}}}
& Scenario-Actor vs Greedy
& $+0.12$
& $[-0.11,+0.35]$
& $3.09 \times 10^{-1}$ \\
& Scenario-Actor vs Fixed $T=0.5$
& $+0.12$
& $[-0.12,+0.36]$
& $3.29 \times 10^{-1}$ \\
& Scenario-Actor vs Fixed $T=0.7$
& $+0.08$
& $[-0.13,+0.29]$
& $4.63 \times 10^{-1}$ \\
& Scenario-Actor vs Fixed $T=1.0$
& $+0.30$
& $[+0.05,+0.55]$
& $1.91 \times 10^{-2}$ \\
& Scenario-Actor vs Fixed $T=1.5$
& $+1.15$
& $[+0.79,+1.51]$
& $3.3 \times 10^{-10}$ \\
& Scenario-Actor vs EDT
& $+1.17$
& $[+0.80,+1.54]$
& $9.03 \times 10^{-10}$ \\
& Scenario-Actor vs Nucleus
& $+0.28$
& $[+0.05,+0.51]$
& $1.86 \times 10^{-2}$ \\
& Scenario-Actor vs Random-$T$
& $+0.20$
& $[+0.01,+0.39]$
& $4.13 \times 10^{-2}$ \\
& Scenario-Actor vs Class-conditioned $T$
& $+0.27$
& $[+0.02,+0.52]$
& $3.21 \times 10^{-2}$ \\
\midrule
\multirow{8}{*}{\textbf{\shortstack[l]{GSM8K\\($\Delta$ Accuracy)}}}
& Scenario-Actor vs Greedy
& $-11.90$
& $[-14.71,-9.02]$
& $6.1 \times 10^{-16}$ \\
& Scenario-Actor vs Fixed $T=0.5$
& $-4.25$
& $[-7.13,-1.36]$
& $4.7 \times 10^{-3}$ \\
& Scenario-Actor vs Fixed $T=0.7$
& $-3.79$
& $[-6.80,-0.80]$
& $1.2 \times 10^{-2}$ \\
& Scenario-Actor vs Fixed $T=1.0$
& $+2.88$
& $[-0.23,+5.99]$
& $7.5 \times 10^{-2}$ \\
& Scenario-Actor vs Fixed $T=1.5$
& $+20.09$
& $[+16.83,+23.35]$
& $2.5 \times 10^{-31}$ \\
& Scenario-Actor vs EDT
& $-3.87$
& $[-6.90,-0.83]$
& $1.5 \times 10^{-2}$ \\
& Scenario-Actor vs Nucleus
& $+3.94$
& $[+0.76,+7.20]$
& $1.7 \times 10^{-2}$ \\
& Scenario-Actor vs Random-$T$
& $-0.53$
& $[-3.79,+2.73]$
& $7.8 \times 10^{-1}$ \\
\bottomrule
\end{tabular}
\caption{Statistical significance testing results on AgentSense, AlpacaEval, MT-Bench, and GSM8K. Differences are computed as Scenario-Actor minus the corresponding baseline.}
\label{tab:statistical_significance}
\end{minipage}
\end{table*}
\FloatBarrier

\begin{table*}[!t]
\section{Scenario Descriptions}
\label{app:scenario-descriptions}
\centering
\setlength{\tabcolsep}{5pt}
\begin{tabular}{@{}p{0.23\textwidth}p{0.70\textwidth}@{}}
\toprule
Scenario Label & Description \\
\midrule
rigorous & Interactions centered on correctness, logical structure, technical precision, or verifiable reasoning. \\
institutional & Interactions organized around explicit roles, responsibilities, procedures, or task objectives. The respondent should help resolve the issue efficiently. \\
formal & Interactions where etiquette, tact, face-saving, and socially careful wording are central. \\
emotional\_support\_neg & Interactions in which the user expresses distress, worry, sadness, frustration, vulnerability, or seeks comfort. \\
adversarial & Interactions involving disagreement, critique, challenge, defense, debate, or pressure to justify a position. \\
emotional\_support\_pos & Interactions in which the user shares positive affect, appreciation, celebration, affection, or seeks a warm affirming response. \\
casual\_conversation & Interactions that are low-stakes, socially light, and not primarily task-, role-, or problem-driven. \\
brainstorming & Interactions that ask for idea generation, alternatives, exploration, or open-ended ideation. \\
playful & Interactions that are primarily humorous, teasing, and entertaining. \\
\bottomrule
\end{tabular}
\caption{Scenario labels and their corresponding communicative descriptions.}
\label{tab:scenario-descriptions}

\section{Layer Sweep and Geometric Diagnostics}
\label{app:layer-sweep}
\centering
\begin{tabular}{lcccccc}
\toprule
Layer & Recall@1 & Effective Rank & tanh Sat. & PCA@95\% & Max Axis Offdiag & Collapse \\
\midrule
16 & 0.770 & 24.41 & 4.33\% & 20 & 0.60 & acceptable \\
20 & 0.798 & 22.80 & 15.42\% & 16 & 0.80 & acceptable \\
24 & 0.739 & 18.21 & 50.70\% & 11 & 0.74 & moderate \\
29 & 0.576 & 10.95 & 88.17\% & 6 & 0.99 & severe \\
30 & 0.518 & 7.08 & 92.24\% & 4 & 0.94 & severe \\
\bottomrule
\end{tabular}
\caption{
Layer sweep diagnostics for scenario geometry. Collapse severity is assessed jointly from tanh saturation, effective rank, and axis off-diagonal correlation. Deeper layers exhibit increasing saturation, rank collapse, and axis degeneracy.
}
\label{tab:layer-sweep-full}

\section{Agreements}
\label{app:agreements}
\centering
\setlength{\tabcolsep}{3pt}
\begin{minipage}[t]{0.48\textwidth}
\centering
\begin{tabular}{lcc}
\hline
Metric & LLM-human & Human-human \\
\hline
Agreement (\%) & 83.66 / 86.27 & 81.37 \\
Cohen's $\kappa$ & 0.82 / 0.85 & 0.79 \\
Macro Precision & 0.84 & 0.83 \\
Macro Recall & 0.84 & 0.80 \\
Macro F1 & 0.83 & 0.81 \\
\hline
\end{tabular}
\caption{LLM annotator validation, values are rounded to two decimal places}
\label{tab:llm-annotator-validation}
\end{minipage}
\hfill
\begin{minipage}[t]{0.48\textwidth}
\centering
\begin{tabular}{lcc}
\hline
Metric & LLM-human & Human-human \\
\hline
AE Agreement (\%) & 86.00 / 86.00 & 80.00 \\
AE Cohen's $\kappa$ & 0.72 / 0.72 & 0.60 \\
MT Pearson $r$ & 0.92 / 0.86 & 0.87 \\
MT Spearman $\rho$ & 0.84 / 0.70 & 0.79 \\
MT MAE & 0.44 / 0.90 & 0.78 \\
AS Agreement (\%) & 80.00 / 72.00 & 76.00 \\
AS Cohen's $\kappa$ & 0.48 / 0.27 & 0.40 \\
\hline
\end{tabular}
\caption{LLM judge validation, values are rounded to two decimal places.}
\label{tab:llm-judge-validation}
\end{minipage}
\end{table*}
\FloatBarrier

\begin{figure*}[ht]
\begin{minipage}{\textwidth}
\section{Source Composition}
\label{app:source-composition}

\centering
\includegraphics[
    width=\textwidth,
    page=1
]{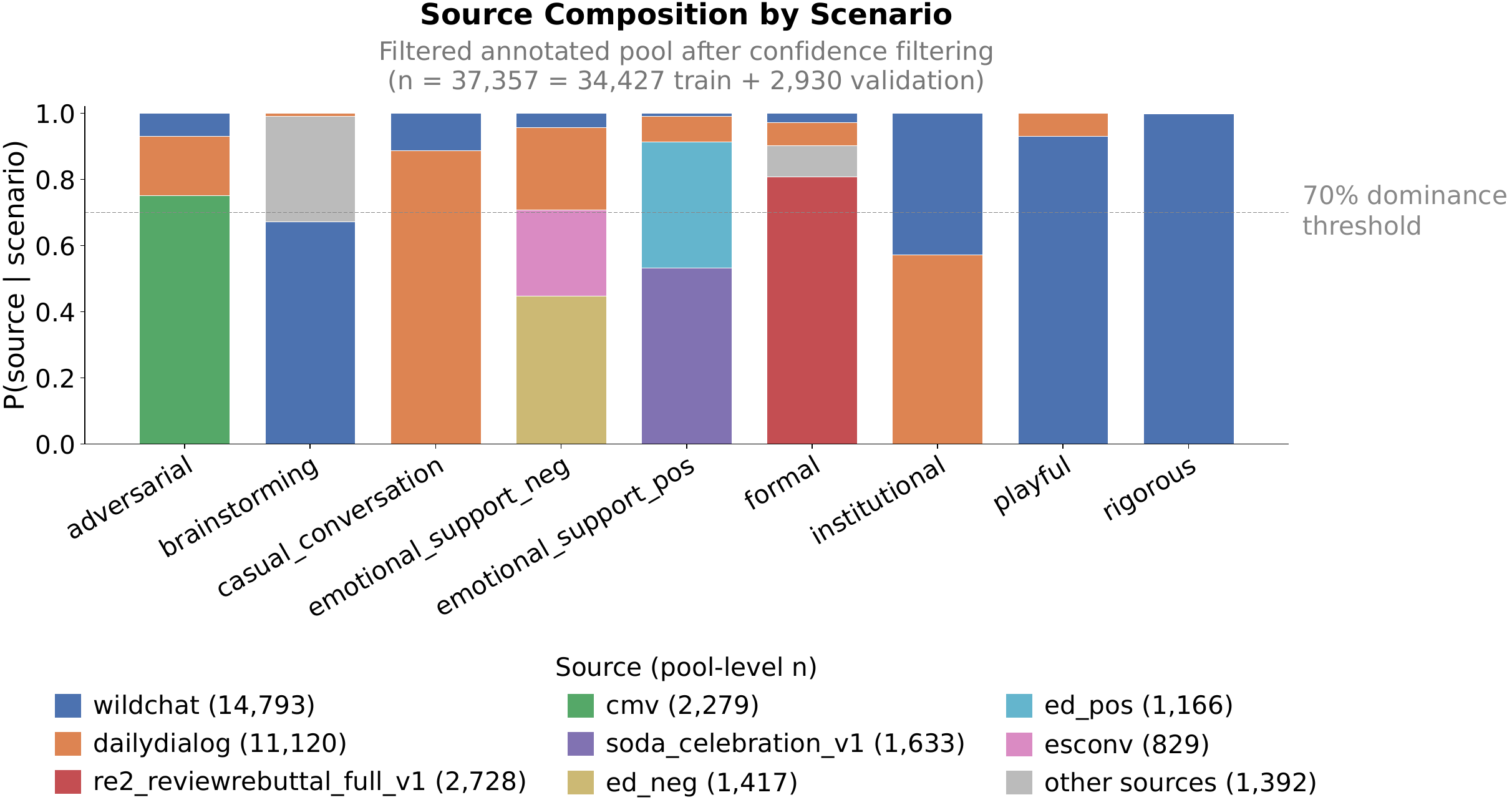}
\caption{Source composition by scenario.}
\label{fig:source-composition}
\end{minipage}
\end{figure*}

\begin{table*}[!t]
\begin{minipage}{\textwidth}
\section{Triplet-Margin Ablation}
\label{app:beta-margin}
\centering
\setlength{\tabcolsep}{7pt}

\begin{tabular}{@{}lccc@{}}
\toprule
\textbf{Geometry metric}
& $\boldsymbol{\beta=0.3}$
& $\boldsymbol{\beta=1.0}$
& $\boldsymbol{\Delta}$ \\
\midrule
Mean pairwise centroid cosine & 0.9496 & 0.7103 & $-0.2393$ \\
Centered centroid cosine & 0.5258 & 0.5057 & $-0.0201$ \\
Centroid effective rank at 95\% & 5 & 6 & $+1$ \\
Euclidean silhouette & 0.1835 & 0.2668 & $+0.0833$ \\
Nearest-centroid agreement & 0.6940 & 0.8877 & $+0.1937$ \\
tanh saturation ($>0.95$) & 0.0938 & 0.0347 & $-0.0591$ \\
\bottomrule
\end{tabular}
\caption{Scenario geometry under two triplet-loss margins.}
\label{tab:beta-margin-geometry}

\vspace{1em}

\begin{tabular}{@{}llccc@{}}
\toprule
\textbf{Benchmark}
& \textbf{Metric}
& $\boldsymbol{\beta=0.3}$
& $\boldsymbol{\beta=1.0}$
& $\boldsymbol{\Delta}$ \\
\midrule
GSM8K & Accuracy (\%) & 65.28 & 66.41 & $+1.13$ \\
AlpacaEval & LC-WR (\%) & 30.03 & 25.54 & $-4.49$ \\
MT-Bench & GPT-4 score & 7.91 & 7.66 & $-0.25$ \\
DailyDialog NextTurn & WR (\%) & 49.70 & 46.94 & $-2.76$ \\
DailyDialog ContextLen & WR (\%) & 67.00 & 62.00 & $-5.00$ \\
\midrule
\multicolumn{5}{@{}l}{\textit{Diagnostic}} \\
GSM8K & Mean selected $T$ & 0.822 & 0.738 & $-0.084$ \\
\bottomrule
\end{tabular}
\caption{Downstream performance under two triplet-loss margins.}
\label{tab:beta-margin-downstream}

\end{minipage}
\end{table*}
\FloatBarrier

\section{Computational Details} 
\label{app:computational-details}

We keep the backbone model frozen and train only the $\operatorname{Scenarize}$ module, which maps the 4096-dimensional Llama hidden states into a 32-dimensional scenario embedding. All experiments run on a single NVIDIA A10G GPU with 24GB memory. The total computational budget, including hidden-state extraction, $\operatorname{Scenarize}$ training, layer sweeps, and evaluation runs, is approximately 125 A10G GPU-hours. This estimate excludes external API inference costs used for GPT-4o annotation and GPT-4-turbo evaluation.

\section{Artifact licenses and terms of use.}
We use Llama-3.1-8B-Instruct, WildChat, DailyDialog, ChangeMyView \citep{tan2016cmv}, SODA \citep{kim2023soda}, EmpatheticDialogues \citep{rashkin2019empathetic}, ESConv \citep{liu2021esconv}, OpenAssistant \citep{kopf2023openassistant}, Re$^2$ \citep{zhang2025re2}, DISAPERE \citep{kennard2022disapere}, OpenReview review text\footnote{\url{https://openreview.net}}, Enron email corpus \citep{klimt2004enron}, GPT-4o, and GPT-4-turbo, and previously proposed decoding baselines in accordance with their respective licenses or terms of use. All datasets are used only for research purposes, and no attempt is made to redistribute restricted model weights or proprietary model outputs beyond aggregate experimental results.

\end{document}